\documentclass[
twocolumn,
]{ceurart}

\usepackage{listings}
\usepackage{siunitx}
\usepackage{subcaption}
\usepackage{hyperref}
\usepackage{cleveref}
\usepackage{combelow}

\usepackage{float}
\usepackage{appendix}

\begin{document}

\copyrightyear{2025}
\copyrightclause{Copyright for this paper by its authors.
  Use permitted under Creative Commons License Attribution 4.0
  International (CC BY 4.0).}

\conference{CLiC-it 2025: Eleventh Italian Conference on Computational Linguistics, September 24 — 26, 2025, Cagliari, Italy 
}

\title{Dissonant ballerinas and crafty carrots: a comparative multi-modal analysis of Italian brain rot}

\author[1,2]{Anca Dinu}[%
orcid=0000-0002-4611-3516,
email=anca.dinu@lls.unibuc.ro
]
\cormark[1]

\author[1,3]{Andra-Maria Florescu}[%
orcid=0009-0007-1949-9867,
email=andra-maria.florescu@s.unibuc.ro
]

\author[1, 3, 4]{Marius Micluța-Câmpeanu}[%
orcid=0009-0002-3197-170X,
email=marius.micluta-campeanu@.s.unibuc.ro
]

\author[1,3]{Stefana-Arina Tăbușcă}[%
orcid=0009-0003-3376-6416,
email=stefana.tabusca@s.unibuc.ro
]

\author[1, 3, 4]{Claudiu Creangă}[%
orcid=0009-0005-5443-5506,
email=claudiu.creanga@s.unibuc.ro
]

\author[1, 4]{Andreiana Mihail}[%
orcid=0009-0003-0640-6319,
email=andreiana.mihail@s.unibuc.ro
]

\address[1]{University of Bucharest,
  Bd. Regina Elisabeta no. 4-12, Bucharest}

\address[2]{Faculty of Foreign Languages and Literatures, Str. Edgar Quinet no. 5-7, Bucharest}

\address[3]{Interdisciplinary School of Doctoral Studies, Bd. Mihail Kogălniceanu no. 36-46, Bucharest}

\address[4]{Faculty of Mathematics and Computer Science, Str. Academiei no. 14, Bucharest}

\cortext[1]{Corresponding author.}

\begin{abstract}
This paper presents a comparative multi-modal analysis of Italian and Romanian brain rot memes, investigating the factors that contribute to its appeal and the linguistic and cultural distinctions between the two versions. To conduct this analysis, we introduce a multi-modal brain rot dataset named CRIB (\textbf{C}ollection of \textbf{R}omanian and \textbf{I}talian \textbf{B}rain rot), a manually curated collection of 240 TikTok videos stratified by language (Italian, Romanian) and popularity, on which we examine textual, acoustic, and visual features. Our findings indicate that popularity is not significantly correlated with textual elements like sentiment, absurdity, or rhyme, or acoustic elements such as vocal features or sentiment of the sound. Instead, in Romanian language, video-level dynamics, specifically faster cutting speeds and a more rapid overall pace, are strong predictors of a video's success. The cross-linguistic analysis reveals significant differences. Italian brain rot is textually more negative, exhibits higher perplexity, and uses more rhyme, while its sound is characterized by higher melodic range and loudness. Romanian audio is spectrally brighter with more erratic pitch variations. 
\end{abstract}


\maketitle

\section{Introduction}

The first recorded use of ‘brain rot’ is in Henry David Thoreau’s book \textit{Walden}, published in in 1854, which criticizes society’s tendency to devalue complex ideas in favor of simple ones, indicating a general decline in mental and intellectual effort: “While England endeavors to cure the potato rot, will not any endeavor to cure the brain-rot – which prevails so much more widely and fatally?” \cite{thoreau2011walden}.

Brain rot was the Oxford word of the year 2024.\footnote{\href{https://corp.oup.com/word-of-the-year/2024/}{Oxford University Press Word of the Year 2024}}
Its primary sense is the supposed deterioration of a person’s mental or intellectual state as the result of over-consumption of low-quality, trivial, or non-challenging online content. Its secondary sense, acquired over the last year, is the online content itself likely to lead to such deterioration.
In particular, the term came to be used in the last months to refer to a certain type of multi-modal short content, intentionally created to be absurd, nonsensical, dissonant and funny, by content developers using generative AI. One of the earliest examples is \textit{Nothing, Forever} in December 2022\footnote{\href{https://news.ycombinator.com/item?id=33833645}{Hacker News post}, 2\textsuperscript{nd} December 2022} (shortly after the launch of ChatGPT), while Italian brain rot is a more recent trend that gained popularity in early 2025. As a side note, short animations like \textit{Skibidi toilet} series or \textit{only in Ohio} memes series are not usually created using generative AI, though they are also considered a form of brain rot.

\begin{figure}
    \centering
    \includegraphics[width=0.5\linewidth]{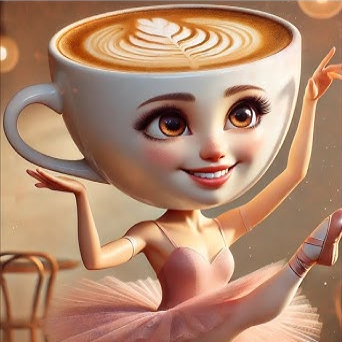}
    \caption{Example of Italian brain rot character \textit{Ballerina Cappuccina}}
    \label{fig:ballerina}
\end{figure}

\begin{figure}
    \centering
    \includegraphics[width=0.5\linewidth]{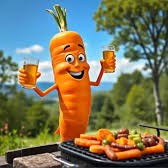}
    \caption{Example of Romanian brain rot character \textit{Morcoveață}}
    \label{fig:Morcoveață}
\end{figure}

The creator of some of the first Italian brain rots, like Ballerina Cappuccina, supposedly a Romanian,\footnote{\href{https://www.nytimes.com/2025/04/30/style/italian-brain-rot-ballerina-cappuccina.html}{New York Times article}, 30\textsuperscript{th} April 2025}\textsuperscript{,}\footnote{\href{https://mindcraftstories.ro/societate/te-mira-ca-in-spatele-italian-brainrot-e-un-roman/}{Interview in Romanian, Mindcraft Stories}, 27\textsuperscript{th} May 2025} describes them as a satiric artistic experiment that both mocks and celebrates pop culture and kitsch. The characters in these creations are childlike, weird, and often grotesque blends of humans, animals, plants and various objects, named with Italian-sounding names, like \textit{Ballerina Cappuccina} in figure \ref{fig:ballerina}. The Italian brain rot phenomenon gained traction especially among Gen Z and Gen Alpha communities by means of social media platforms like TikTok and Instagram and has quickly spread to other languages, including Romanian, with characters such as \textit{Morcoveață} in figure \ref{fig:Morcoveață}, a human-like carrot character adapted from Romanian nursery rhymes.

A common trait of brain rot is the uncertain or lack of authorship. As Roland Barthes argued in \textit{The Death of the Author} \cite{barthes1977death}, removing the author takes away a traditional source of authority and shifts the focus to the reader and their interpretation. In a similar way, brain rot content is mostly anonymous or pseudonymous, remixed and layered, reflecting an anarchic and collective form of creation. 

Moreover, brain rot has many common traits with the Dadaism movement, which was on the same path of anti-art and anti-meaning, it mocked the art rules and traditions, and it embraced nonsense and absurdity. It celebrated chaos and irrationality, at the same time being psychedelic in the sense of artistic production from the 70s and 80s  when using vivid colors, fractals, surreal images, and neon gradients. Both forms of art were born in an age of post truth and information overload.

Finally, the brain rots' original purpose seems to be primarily amusement, but some of them were also used for manipulation, commercial purposes or even political propaganda. 


This study aims to explore whether brain rot manifests differently across cultures and languages. We chose to begin with Italian brain rot, as it represents one of the most visible and influential starting points for this phenomenon. The decision to compare it with Romanian brain rot is based on both similar and contrasting historical, linguistic, and cultural characteristics. The main common grounds between the two languages and cultures are that they both belong to the Romance language family, and that they have experienced authoritarian regimes in the 20th century that shaped their collective imagination, creative approaches and imagery. The differences between the two lie in their geopolitical contexts, religious traditions (Catholic and Greek Orthodox), and features of cultural production, which influence the tone, style, and content of their digital aesthetics. At the same time, both Italian and Romanian online cultures remain underrepresented in cultural and computational studies, which represents an opportunity to examine how the upper similarities and differences are reflected in brain rot popular manifestation.

The main research questions are: (1) What makes a brain rot go viral (besides algorithm recommendation/manipulation or pure chance)? and (2) Are there any cultural or language differences between the form or content of brain rots in these two languages?





\section{Related Work}

Numerous studies are currently interested in understanding what are the effects of digital content over-consumption. Most of them focus on psychological, neuro-biological, or meta-analytical perspectives \cite{brainsci15030283,Satici2023}. 

 This type of digital content certainly breaks conventional norms of art, narrative, and symbolism, aligning well with both surrealism and absurdism, while also embracing the chaos and disjuncture of postmodernity. The chimera-like mixing of humans, animals, plants, and objects is not a new phenomenon, since anthropomorphism can be traced back from Egyptian, Greek, and Roman antiquity, going through Medieval art, all the way to the surrealist movement at the beginning of the 20th century (Max Ernst, Salvador Dali, or René Magritte). 

To the best of our knowledge, no computational research has been conducted specifically on Italian brain rot. 
However, internet culture, including memes and viral content received a significant amount of attention.  The detection of meme toxicity was investigated in \cite{pandiani2024toxicmemessurveycomputational}. Multimodal sentiment analysis was conducted by integrating text, image, and audio for improving sentiment detection from vlogs, spoken reviews, and human-machine interactions in \cite{SOLEYMANI20173}.
A comprehensive survey which categorizes advances in multimodal sentiment analysis can be found in \cite{GANDHI2023424}.
\cite{kiela2021hatefulmemeschallengedetecting} introduced a new benchmark for detecting hate speech from multimodal memes. \cite{lin2024explainableharmfulmemedetection} combined LLM-generated debates and fine-tuned judge models to detect harmful memes with improved interpretability and performance. 
\cite{bloem2025clusteringinternetmemestemplate} proposed a template-based approach for meme clustering by employing multi-dimensional similarity features.

\section{Data}
The data set was constructed manually from TikTok videos by searching for candidate examples through various methods: direct search queries, tags, the \textit{discover} feature, trending pages, compilation and analysis of video clips, related videos (\textit{You may like}), and the \textit{For You} recommendations. This process cannot be reliably automated due to misleading tags being frequently used to influence the recommendation algorithm.

The extracted samples are stratified across two dimensions: language (Italian and Romanian) and popularity (popular and unpopular). The dataset is well balanced: 120 brain rots per language, with 60 viral examples and 60 less viewed examples, the typical threshold between them being 100k views or at least 10k likes.


Given that we are interested in all aspects of communication, especially creative language use, we filtered out posts with extremely low lexical diversity, as well as re-uploads, translations, and repetitive songs. 

Moreover, some of the tools used to generate these videos can be traced by watermarks, although we note that some users specifically crop the content or blur such indicators. In alphabetical order, clips have been created or adjusted with: \href{https://capcut.com/}{CapCut}, \href{https://chatgpt.com/}{ChatGPT}, \href{https://hailuoai.video/}{Hailuo AI}, \href{https://klingai.com/}{Kling AI} (version 1.6), \href{https://app.pixverse.ai/}{PixVerse.ai}, \href{https://runwayml.com/}{Runway}, \href{https://veed.io/}{VEED}. TikTok also offers its own tools for video creation.


We collected four subcategories of brain rots for each language. The general category in each language includes the notorious Italian brain rot characters with local adaptations (with 60 examples per language). For Italian, the \textit{skeleton} category consists of 20 videos with a poignant tone. The \textit{Matteo} category comprises 20 memes that exhibit more positive attitudes, and the \textit{Politicians and Celebrities} class consists of 20 political satire videos. For Romanian, we followed the same count and structure. We have selected \textit{schelet}, the corresponding category of Italian \textit{schletro}, \textit{Regina brain rot} (brain rot queen), featuring longer stories, and conversely \textit{Morcoveață}, consisting of very short clips.
Throughout this process, we have extracted a total of 240 videos with subtitles and metadata.\footnote{Using \href{https://github.com/yt-dlp/yt-dlp}{yt-dlp} version 2025.04.30.} We name the dataset CRIB (\textbf{C}ollection of \textbf{R}omanian and \textbf{I}talian \textbf{B}rain rot).

\section{Text}

\subsection{Sentiment Analysis of text}

To obtain sentiment analysis scores for the textual transcripts of the brain rots, we employed \href{https://huggingface.co/cardiffnlp/twitter-xlm-roberta-base-sentiment}{cardiffnlp/twitter-xlm-roberta-base-sentiment} pre-trained model \cite{barbieri-etal-2022-xlm}, which returns a percentage of negative, neutral, and positive sentiments associated with each text in both languages, for optimal comparison. We used ChatGPT4o for coding assistance and Python in Google Colab to obtain the graphical illustrations and to perform statistical tests.

\begin{figure}[p] 
  \centering
\begin{subfigure}[t]{0.45\textwidth}
  \includegraphics[width=\textwidth]{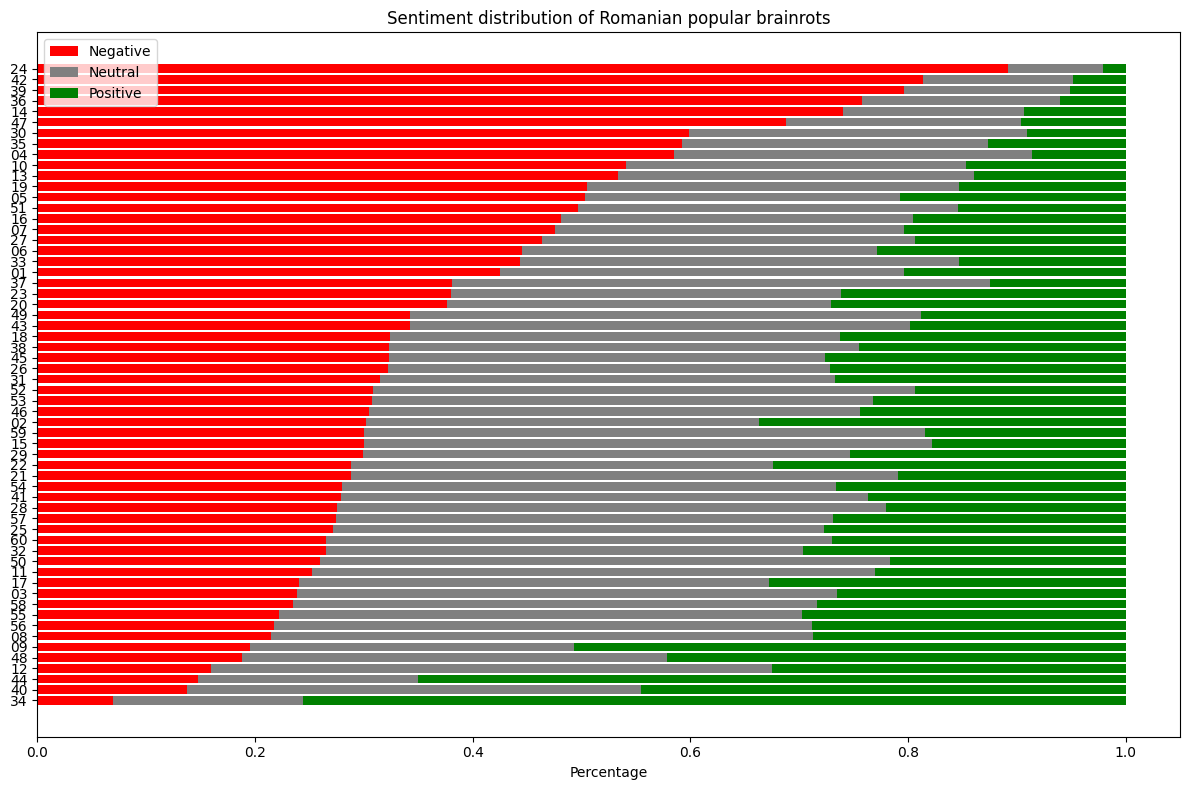}
  \caption{Sentiment scores for Romanian popular brain rots}
  \label{fig:sentimentTextRoPopular}
\end{subfigure}

\begin{subfigure}[t]{0.45\textwidth}
  \includegraphics[width=\textwidth]{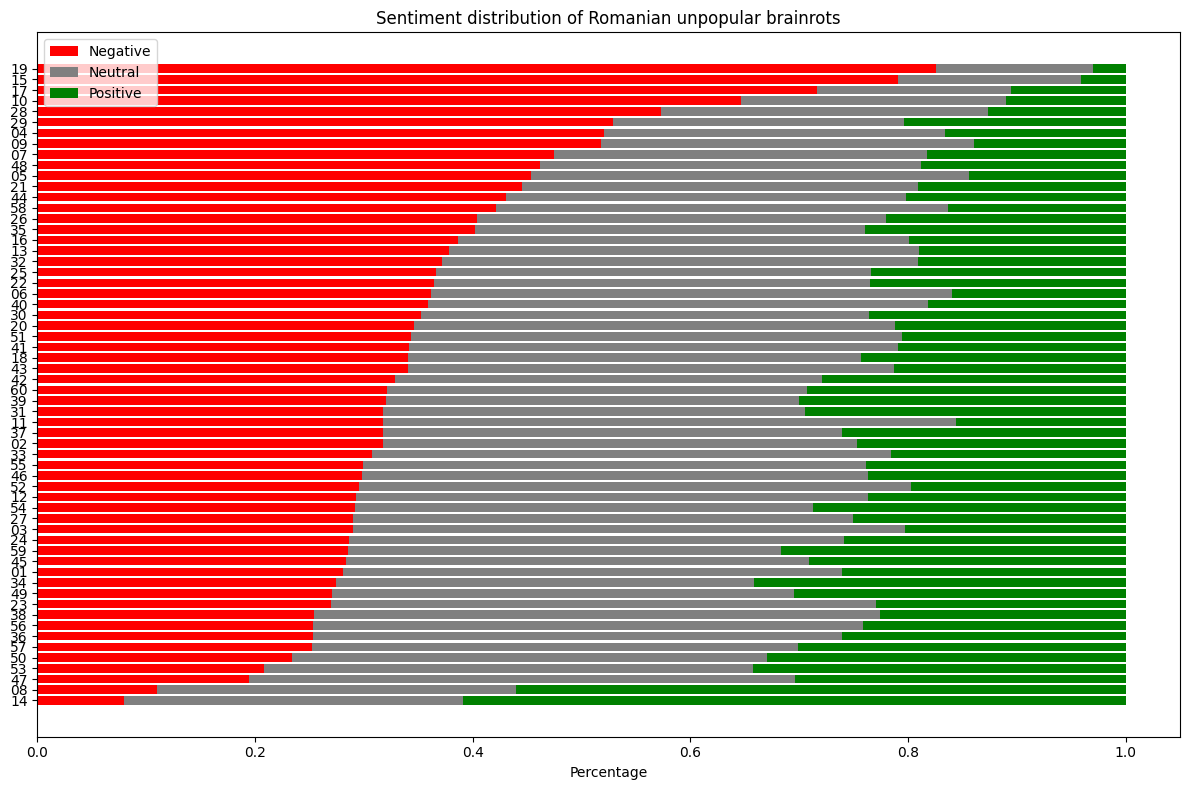}
  \caption{Sentiment scores for Romanian unpopular brain rots}
  \label{fig:sentimentTextRoUnpopular}
\end{subfigure}

\begin{subfigure}[t]{0.45\textwidth}
  \includegraphics[width=\textwidth]{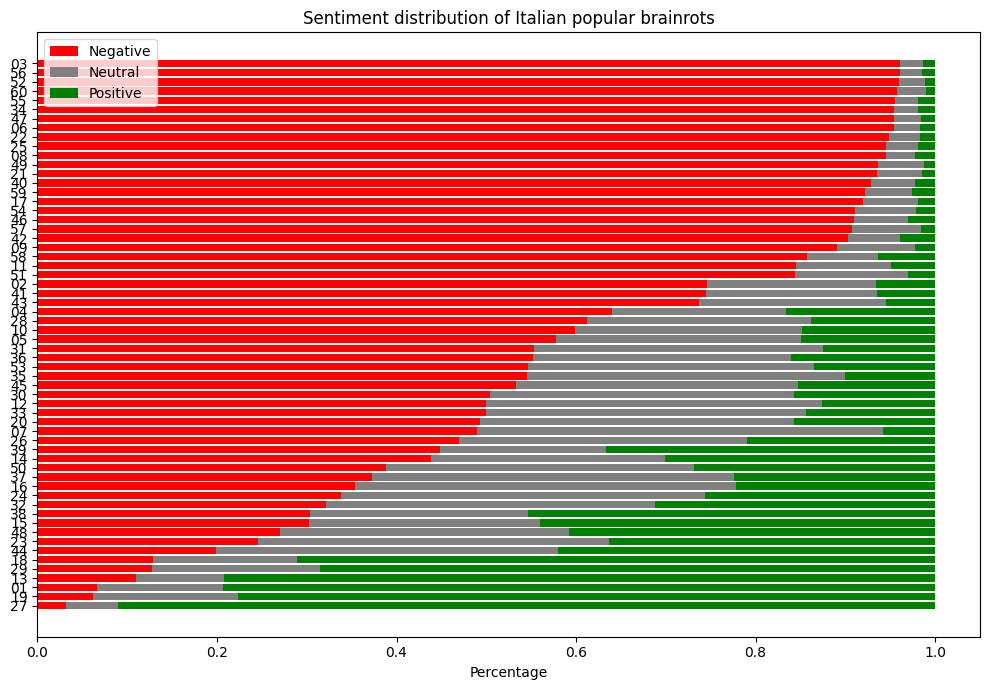}
  \caption{Sentiment scores for Italian popular brain rots}
  \label{fig:sentimentTextItPopular}
\end{subfigure}

\begin{subfigure}[t]{0.45\textwidth}
  \includegraphics[width=\textwidth]{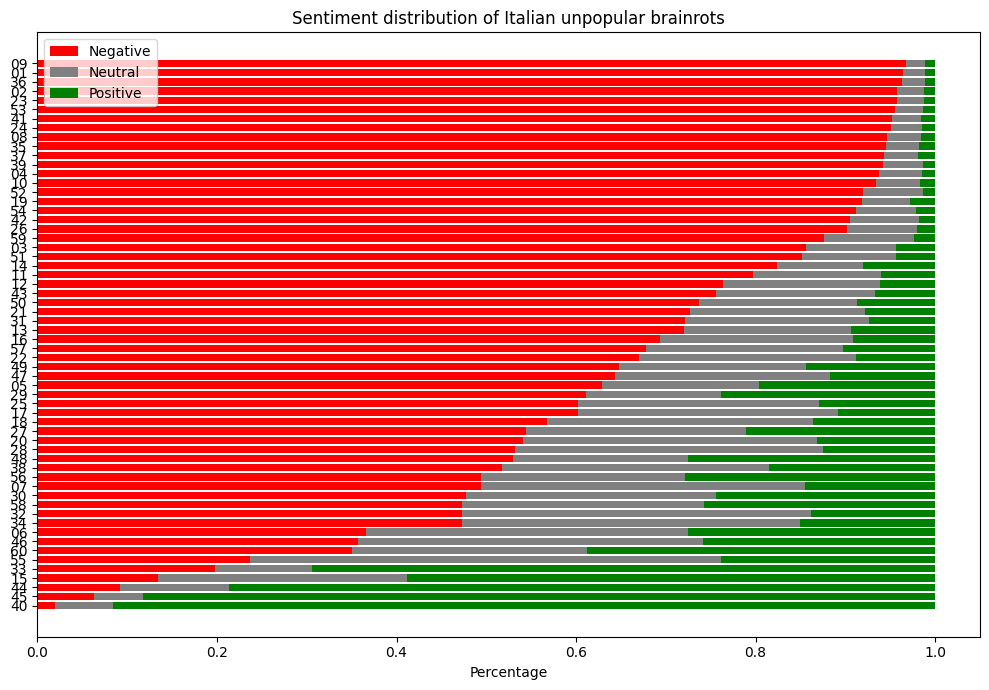}
  \caption{Sentiment scores for Italian unpopular brain rots}
  \label{fig:sentimentTextItUnpopular}
\end{subfigure}
\caption{Sentiment scores (Negative, Neutral, Positive) for each language and popularity group of the brain rot texts}
  \label{fig:allSentimentPlots}

\end{figure}

Overall, for all brain rots, the negative sentiment was predominant. There were minimal variations in sentiments in popular versus unpopular brain rots in both languages, as we can see in figures \ref{fig:sentimentTextRoPopular} and \ref{fig:sentimentTextRoUnpopular} for Romanian and in figures \ref{fig:sentimentTextItPopular} and \ref{fig:sentimentTextItUnpopular} for Italian. However, the Italian ones contained more negative sentiments than their Romanian counterpart.

We tested for statistical significance of these findings and the results are listed in table \ref{tab:statsSentimentsText} from the Appendix.

For Romanian differences of positive, neutral, and negative sentiments between popular and unpopular texts, we performed a multivariate analysis of variance (MANOVA). The results indicate no statistically significant multivariate effect. To further investigate potential differences at the level of individual sentiments, given potential concerns about normality and homogeneity of variance,  Mann-Whitney U tests were performed for each emotional score. The results confirmed that there are no statistically significant differences. 

The same tests on the Italian brain rots obtained the same results: no statistical differences between the sentiments of popular and unpopular texts. 

We further performed, for each language, the same statistical tests only on the \textit{general} category of brain rots, which varies less than the combination of all four brain rot categories (\textit{general}, \textit{regina brain rot}, \textit{morcoveață}, and \textit{schelet} for Romanian, and \textit{general}, \textit{matteo}, \textit{politicians and celebrities}, \textit{scheletro} for Italian) to test the differences between popular and unpopular categories. The results showed again no significant differences.

Finally, a multivariate analysis of variance (MANOVA) was conducted to examine whether the distribution of emotional sentiment scores (positive, neutral, and negative) differed between Romanian and Italian brain rots. This time, results were statistically significant, indicating a robust multivariate effect of language on sentiment composition. To further explore individual sentiment differences between languages, Mann-Whitney U tests were applied separately for each sentiment category. Results revealed that positive sentiment was significantly higher in Romanian brain rots (p < 0.0001; mean: 0.377 vs. 0.159), that neutral sentiment was also significantly higher in Romanian memes (p < 0.0001; mean: 0.385 vs. 0.171), and that negative sentiment was significantly higher in Italian ones (p < 0.0001; mean: 0.670 vs. 0.238).

\subsection{Semantic Similarity and Perplexity}

To measure the degree of ``absurdity'' or ``unpredictability'' of the brain rots, we employed two complementary measures: semantic similarity and perplexity. Semantic similarity was computed for a given text of a brain rot as the average pairwise cosine similarity between all word embeddings obtained with the sentence transformer \textit{\href{https://huggingface.co/sentence-transformers/paraphrase-multilingual-MiniLM-L12-v2}{paraphrase-multilingual-MiniLM-L12-v2}} \cite{reimers-gurevych-2019-sentence}. This reflects how semantically cohesive the vocabulary is — higher values indicate that the words tend to belong to similar semantic fields or contexts, suggesting internal consistency or coherence, while lower values indicate some incoherence or inconsistencies. Perplexity quantifies how unpredictable a text is from the perspective of a pre-trained language model (GPT-2 \cite{radford2019language}). Lower perplexity values indicate that the model finds the sequence more predictable and fluent, while higher values suggest syntactic or lexical irregularities, or content that deviates from typical language patterns.


For both semantic similarity and perplexity scores, we tested the differences in means between popular and unpopular brain rots per language and we also tested the difference in mean between the two languages. We used one-way ANOVA and we also conducted non-parametric Mann-Whitney U tests. The results of the statistical tests are summarized in table \ref{tab:stats_combinedSemanticSimilarityAndPerplexity} from the Appendix.

For Romanian brain rots, we evaluated whether the two metrics, semantic similarity and perplexity, differ significantly between brain rot texts categorized as popular versus unpopular. The results of the ANOVA test revealed no statistically significant differences.

The Mann-Whitney tests results also show that there is no significant difference between popular and unpopular brain rot texts for neither semantic similarity or perplexity scores. However, the p-value for semantic similarity was very close to the significance threshold (0.068), indicating a slight preference for more irregular/nonstandard language in popular brain rot texts (M = 168.59 for popular, M = 159.22 for unpopular).

We also performed statistical tests on the Romanian \textit{general} subcategory of brain rots that varies less than the whole Romanian dataset. The ANOVA results confirmed the ones obtained with the the same statistical tests on all the data, with similar p-values that did not surpass the 0.05 threshold. This time, the semantic similarity p-value of the Mann-Whitney test revealed a statistically significant difference in semantic similarity (p < 0.05), suggesting that popular brain rots are slightly more semantically coherent (with a difference in mean of 0.002) than unpopular brain rots. 

For Italian, the statistical test results suggest that neither semantic similarity, nor perplexity differs significantly across the popular and unpopular brain rot texts. As in the case of Romanian, we also tested the statistical significance of the differences between popular and unpopular texts w.r.t. semantic similarity and perplexity scores, for Italian \textit{general} category of brain rots, but the results were not statistically significant.

Cross-linguistically, we tested whether semantic similarity and perplexity differ significantly between Romanian and Italian brain rot texts. The difference in mean semantic similarity score between Italian (M = 0.59) and Romanian  (M = 0.58) was very small. The perplexity scores that reflect language uncertainties or language inconsistencies were substantially higher for Italian brain rot texts (M = 218.74) than for Romanian ones (M = 163.91). While one-way ANOVA test revealed no significant difference in semantic similarity between Italian and Romanian, for perplexity it yielded a statistically significant effect of the language category with a p-value < 0.001, suggesting that the Italian brain rot texts are indeed more language-unpredictable than their Romanian counterparts. The Mann-Whitney tests confirmed the ANOVA results showing no significance for semantic similarity (p = 0.06), and a highly significant difference between the two groups for perplexity (p < 0.0001).

\subsection{Rhyme}

We estimated the rhyme density by the following methodology. We employed a computational method based on sub-string similarity at word endings in two and three letters. Since the speech to text automatic transcription did not identify correctly the verses with end-line, we checked for all rhyme pairs locally, within a distance of two verses. The rhyme coefficient was calculated as the ratio between the total number of distinct, non-adjacent word pairs that share the same suffix of 2 or 3 letters and the total number of words in the text.


We tested the differences in rhyme scores of popular versus unpopular brain rot texts for both languages, and also the differences in rhyme scores between the two languages with Mann-Whitney tests. The results are shown in table \ref{tab:stats_rhyme} from the Appendix.

The difference of the means between Romanian popular (M = 0.19) and unpopular (M = 0.16)  brain rots suggests a slight preference towards more rhymed ones in the popular group. However, for our sample of 60 (30 popular and 30 unpopular brain rots), this difference is not statistically significant (p = 0.135 $\geq$ 0.05).

For Italian, there is also no statistical difference between the rhyme coefficient of the popular and unpopular groups, since their means are very close (0.215 vs. 0.228) and the p-value is 0.6880 (>0.05).

We also compared Romanian and Italian rhyme coefficients (120 each). The mean for Italian rhyme coefficient is 0.221, higher than the mean for Romanian which is 0.177. This time, the Mann-Whitney test returned the p-value of 0.0001 (<0.05), which means that Italian brain rots do rhyme more than their Romanian counterparts. 

\section{Sound}

\subsection{Audio Processing and Separation}

We developed a multi‐stage audio pipeline to isolate and characterize the vocal component of Italian and Romanian brain rot, with the goal of quantifying attributes that may influence their popularity, as well as perform comparisons between the two languages. First, each video file was converted to a high‐resolution WAV format and processed with a state‐of‐the‐art source‐separation model (Demucs) \cite{rouard2023hybrid} to obtain separate vocal and music stems. Vocal tracks were then normalized for consistent loudness, ensuring that subsequent analyses would not be affected by variations in recording level.

\subsection{Speech Features}

After isolating the vocal stems, we computed a comprehensive set of acoustic descriptors. Pitch contours were extracted via \textit{\href{https://librosa.org/}{librosa}}’s \cite{mcfee_2025_15006942} \texttt{pyin} algorithm, yielding mean F0, F0 variance, 95th–5th percentile range, and slope‐entropy to quantify melodic movement. On 25 ms/10 ms‐hop frames, we computed Mel‐frequency cepstral coefficients (MFCCs) 1–13 (means and variances), spectral centroid, bandwidth, roll-off, zero‐crossing rate, and spectral‐flux (means and variances) to capture brightness, noisiness, and timbral dynamics. Rhythmic patterns were quantified by onset detection—calculating syllable‐rate and pause‐duration statistics—and by RMS (root mean square) energy envelopes (temporal centroid, mean, and variance). Each clip’s features formed one row in a master data matrix.

We then compared these matrices in two ways: within each language (popular vs. unpopular memes) and across languages (Italian vs. Romanian). All continuous features were tested with Mann–Whitney U and corrected for multiple comparisons using the Benjamini–Hochberg procedure at q < 0.05 \cite{benjamini-95-controlling}.

After correction, for the analyses of the popular vs. unpopular groups within each language, no features remained significant, likely due to modest sample sizes and the number of comparisons. Reporting raw p < 0.05 findings (visible in Table \ref{tab:pop_unpop_features} from the Appendix) highlights, though, that in the Italian dataset, popular brain rots exhibited significantly greater fundamental‐frequency variance (var\_f0) and a wider pitch range (range\_f0) than their unpopular counterparts. These results suggest that a more expressive, melodic content, characterized by larger and more varied pitch intervals, is associated with higher popularity in Italian clips. In the Romanian corpus, popular audios differed from unpopular ones in three respects: they featured a marginally faster onset rate (syllable\_rate), a more stable mid-high spectral texture (mfcc\_9\_var), and a darker high-frequency timbre (mfcc\_11\_mean). These patterns imply that a slightly quicker rhythm, smoother structure in the mid-high band, and less pronounced very high frequencies tend to co‐occur with popularity. None of these comparisons survived false discovery rate (FDR) correction at q < 0.05, reflecting exploratory sample size considerations.

In addition to our broad comparisons, we also examined popular versus unpopular clips within each thematic category—namely \textit{general}, \textit{scheletro}, \textit{matteo}, and \textit{politicians and celebrities} in Italian and \textit{general}, \textit{schelet}, \textit{morcoveață}, and \textit{regina brain rot} in Romanian. After applying Benjamini–Hochberg correction, only the Romanian \textit{general} subset yielded features that remained significant. In those 60 clips, one clear acoustic hallmark of popularity is a smoother timbre in the mid‐to‐high frequencies. Specifically, popular memes have a noticeably lower mfcc\_9\_var (181 vs. 232 in unpopular clips), meaning the shape of the spectrum around 3–4 kHz stays more consistent rather than flitting up and down. This steadiness makes the audio feel more even and less ``choppy''. There is also a trend toward lower mfcc\_11\_var (170 vs. 193), which means fewer abrupt jumps in the very high frequencies (around 5–6 kHz), so sibilance and hiss are kept under tighter control. These findings suggest that Romanian \textit{general} brain rots become more engaging when their soundtrack maintains a steady, coherent tone.

When we examine the full set of 120 Italian and 120 Romanian brain rots side by side (as can be seen in Table \ref{tab:italian_vs_romanian} from the Appendix), a clear pattern of prosodic contrast emerges. Italian vocals move through a broader and more dynamic pitch range, peaking and dipping over a span that is markedly larger than in the Romanian samples, which makes them feel more melodically engaging. Romanian clips, on the other hand, swing their pitch contours in a less predictable fashion, with higher entropy of slope suggesting sudden twists in intonation that can come across as more spontaneous or volatile. In the spectral domain, Romanian brain rot tracks push energy into the upper frequencies: their average spectral centroids, bandwidths, and roll-off points all sit higher than in Italian. Yet these high‐frequency bands in Romanian audio are less stable over time, fluctuating more from moment to moment and lending a grainier, more restless timbral texture. Italian clips display a darker, more subdued high‐end, but they deliver their muted tones with greater consistency and, at the same time, add sharp bursts of change, especially in those same upper bands, so that the audio doesn't feel monotone. They also sound uniformly louder, amplifying their broad, dramatic pitch gestures and deep spectral shifts, whereas Romanian tracks embrace a leaner, quieter tone. These findings point to two distinct audio “signatures” that likely reflect both the underlying voice synthesis models and the cultural styles of meme production in each language.

\subsection{Sentiment Analysis of sound}

\begin{figure}[p]
  \centering
  \begin{subfigure}[t]{0.45\textwidth}
    \includegraphics[width=\textwidth]{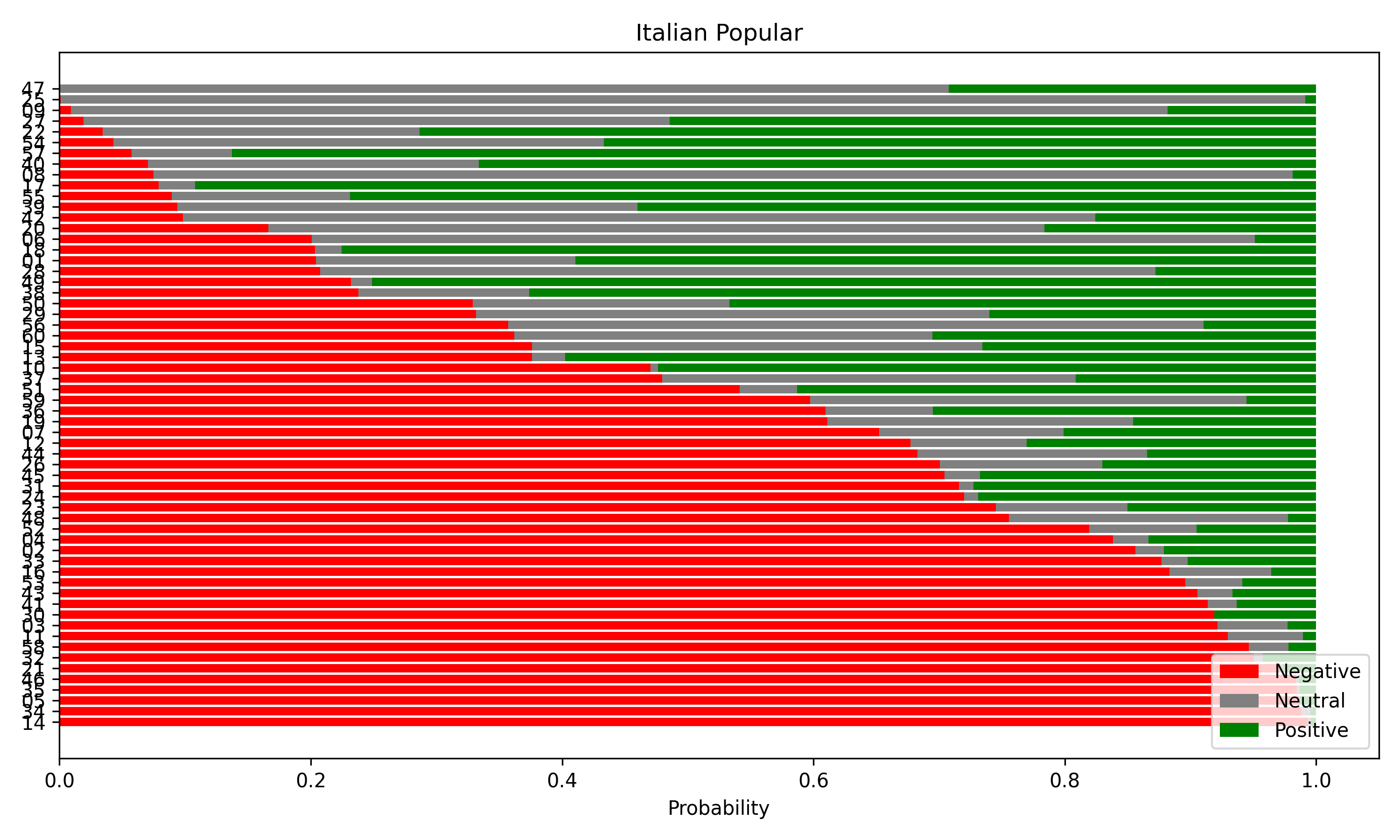}
    \caption{Italian Popular Brain Rot Audios}
    \label{fig:sentItPop}
  \end{subfigure}%
  \hfill
  \begin{subfigure}[t]{0.45\textwidth}
    \includegraphics[width=\textwidth]{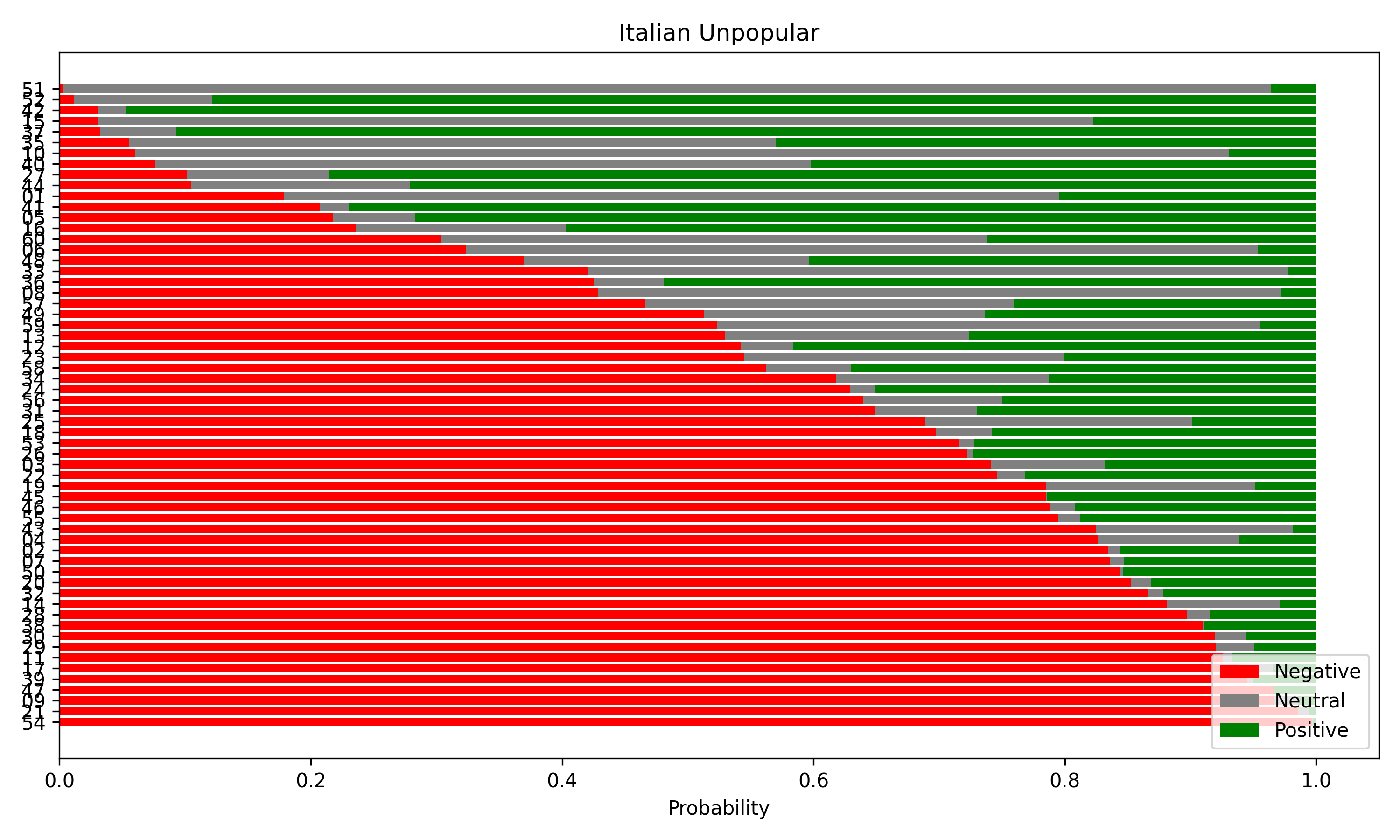}
    \caption{Italian Unpopular Brain Rot Audios}
    \label{fig:sentItUnpop}
  \end{subfigure}

  \vspace{1em}

  \begin{subfigure}[t]{0.45\textwidth}
    \includegraphics[width=\textwidth]{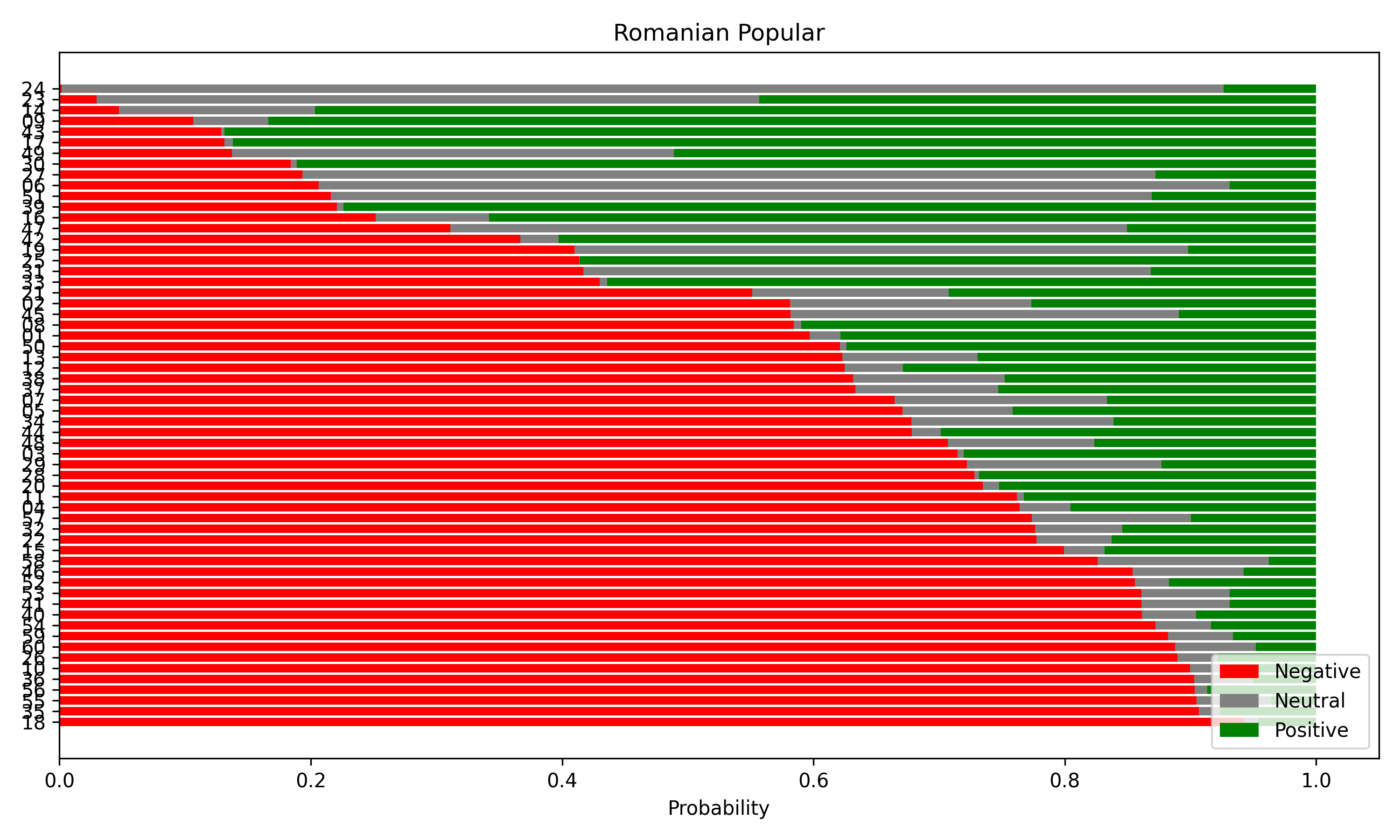}
    \caption{Romanian Popular Brain Rot Audios}
    \label{fig:sentRoPop}
  \end{subfigure}%
  \hfill
  \begin{subfigure}[t]{0.45\textwidth}
    \includegraphics[width=\textwidth]{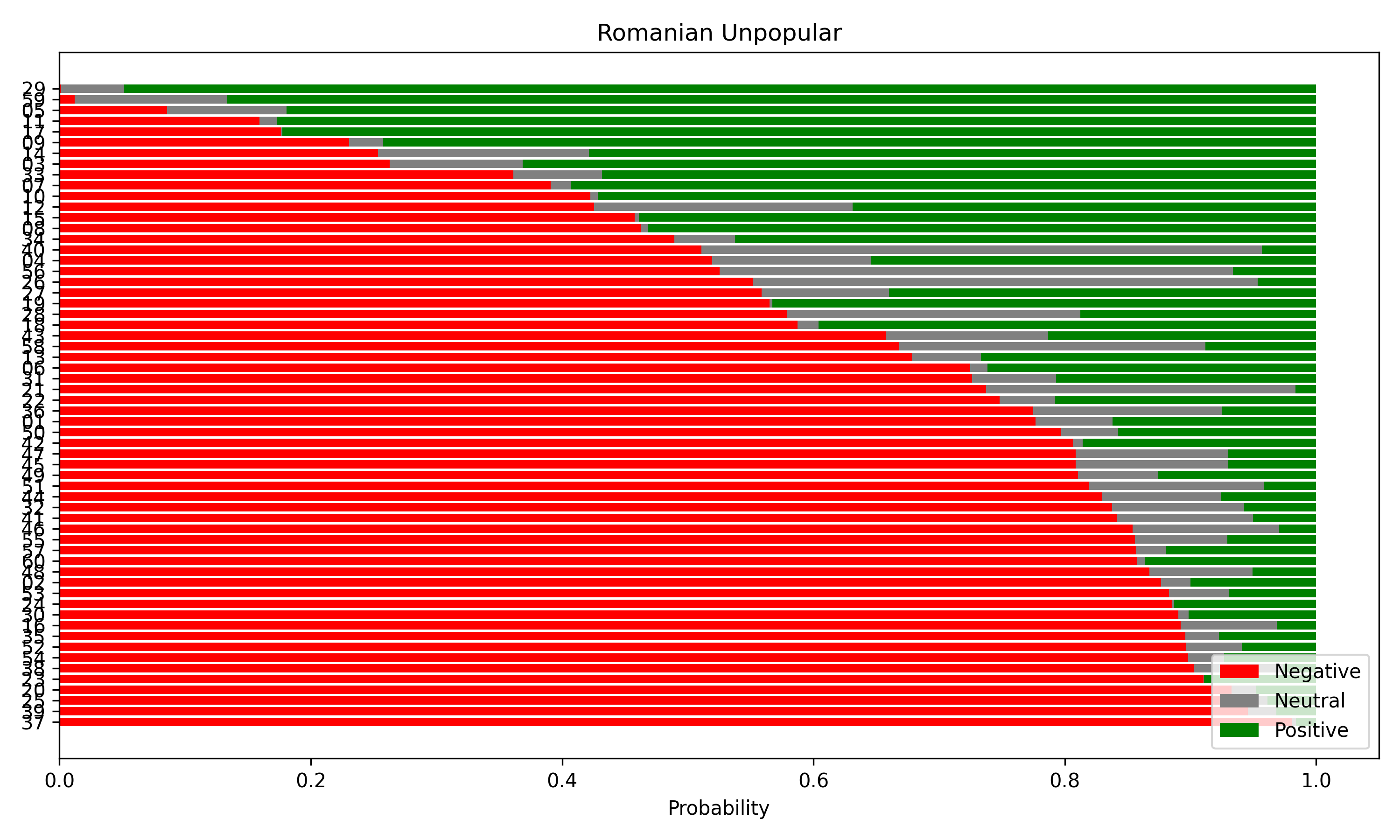}
    \caption{Romanian Unpopular Brain Rot Audios}
    \label{fig:sentRoUnpop}
  \end{subfigure}

  \caption{Continuous‐sentiment distributions (Negative, Neutral, Positive) for each language and popularity group of the brain rot audios}
  \label{fig:allSentimentPlots}
\end{figure}

Full‐audio (voice + music) sentiment was assessed using the pretrained speech‐emotion classifier \href{https://huggingface.co/superb/wav2vec2-base-superb-er}{superb/wav2vec2-base-superb-er}
\cite{yang2021superb}. It categorizes short audio clips into four emotion labels (\textit{angry}, \textit{happy}, \textit{neutral}, \textit{sad}). Each brain rot’s waveform was resampled to 16 kHz and fed into the model, after which we captured the full probability distribution over three coarse categories, Negative (-1), Neutral (0), and Positive (+1). These labels were mapped from the original model outputs as such:  Negative (-1) for any emotion other than \textit{happy} or \textit{neutral} (i.e. \textit{angry} and \textit{sad}), Neutral (0) for labels predicted as \textit{neutral}, and Positive (+1) for labels predicted as \textit{happy}. This mapping yields a better comparison point for our study, in alignment with the sentiment analysis performed for the brain rots' texts. Separate sentiment files were generated for four groups: Italian‐Popular, Italian‐Unpopular, Romanian‐Popular, and Romanian‐Unpopular. The distributions for each audio in the corresponding groups is shown in Figure \ref{fig:allSentimentPlots}.

When we treat the three sentiment probabilities (Negative, Neutral, and Positive) as a multivariate outcome, neither Italian nor Romanian shows a significant difference between popular and unpopular brain rot audios. A MANOVA on the Italian data yields Wilks’ $\lambda$ = 0.992 (F(3,116) = 0.30, p = 0.826), and the same test on Romanian returns Wilks’ $\lambda$ = 0.966 (F(3,116) = 1.37, p = 0.257).

However, comparing Italian versus Romanian clips reveals a modest but statistically significant language effect on the combined sentiment vector (Wilks’ $\lambda$ = 0.965, F(3,236) = 2.87, p = 0.037). In other words, the full‐audio sentiment differs more by language than by popularity within a language.

We also used Mann–Whitney tests to look at each sentiment dimension in isolation and to account for potential concerns about normality and homogeneity of variance. Comparing Italian against Romanian across all 240 clips, the sentiment score distributions show however no robust separation: Negative (U=6,536; p=0.217), Neutral (U=8,088; p=0.099), and Positive (U=6,878; p=0.550) all lie above the conventional 0.05 threshold. The Neutral score comes closest (p $\approx$0.10), hinting at a possible tendency for Italian memes to register slightly higher neutral‐vibe probabilities than Romanian ones, but this trend remains statistically inconclusive.

\section{Video}

To analyze the visual content of the 240 videos, we employed the \href{https://storage.googleapis.com/model-cards/documents/gemini-2.5-flash-preview.pdf}{Gemini Flash 2.5} multimodal model. Each video was individually processed by the model with a prompt instructing it to perform a visual-only analysis and extract a series of predefined attributes. The model returned its analysis for each video as a structured JSON object, which allowed for the systematic collection of data for our study. In an initial test, the model demonstrated a notable capability in discerning the language of origin from visual cues alone, achieving an accuracy of 71.67\% (Romanian vs. Italian) by identifying culturally specific items (e.g., the Carpathian Mountains). However, it struggled significantly to predict a video's popularity, achieving an accuracy of only 47.92\% (random chance). This initial finding suggests that a video's success is not determined by straightforward, immediately classifiable visual markers of appeal. A deeper statistical analysis was therefore performed to uncover more subtle visual attributes that may correlate with popularity (\autoref{table:video_analysis_full} in the Appendix).

\begin{figure}[t]\vspace*{3pt}
\centerline{\includegraphics[width=1\linewidth]{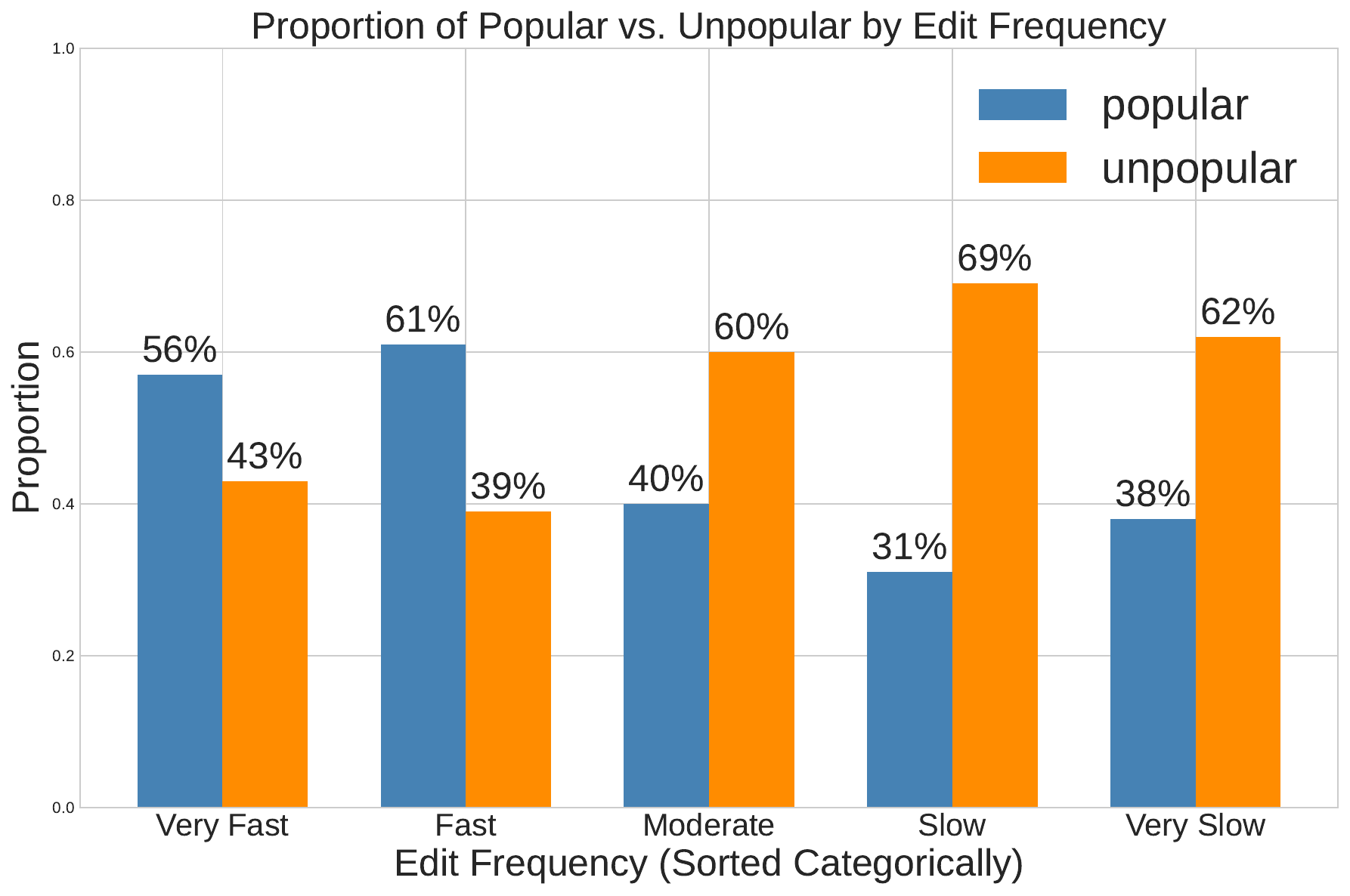}}
\caption{Faster video editing speeds show a stronger correlation with popularity than slower editing speeds}
\label{fig:edit-frequency}
\end{figure}

A key finding emerged from the analysis of the videos' dynamic properties, specifically the rate of  the shot transitions. The data indicates a clear and statistically significant tendency for popular videos to feature a much faster cutting speed. Videos categorized with \textit{Very Fast} or \textit{Fast} transitions were substantially more prevalent among popular content, as it can be seen in figure \ref{fig:edit-frequency}. This observation is supported by a positive Pearson correlation of 0.1712 between cutting speed and popularity, which was found to be statistically significant (p=0.0231). This suggests that as the frequency of cuts increases, so does the likelihood of a video being popular, pointing to a dynamic, high-energy visual style as a key component of audience engagement.

Further reinforcing the importance of dynamism, the overall pacing of the videos also proved to be a significant differentiator. A statistically significant difference was found in the distribution of pacing levels between popular and unpopular videos, as confirmed by a Mann-Whitney U Test (p=0.0492). Popular videos were most frequently described as having a \textit{Fast} overall pace. Moreover, when the correlation between pacing and popularity was analyzed by language, a significant difference was observed. For the Romanian videos, we found a positive Pearson correlation of 0.2791, indicating that as overall pacing increases, videos in Romanian tend to be more popular. In stark contrast, the Italian videos showed a negligible correlation of 0.0047, revealing virtually no linear relationship between pacing and popularity. This language-specific finding suggests that the overall positive trend observed in the combined dataset is almost exclusively driven by the Romanian-language content. While the perceived tempo is an important factor, its influence on popularity appears to be culturally moderated. 

In contrast to the clear influence of dynamism, thematic elements such as absurdity and narrative structure showed no significant correlation with video popularity. Although it was hypothesized that surreal, chaotic, or illogical content might be a good indicator of popularity, the analysis did not bear this out. Perceived absurdity levels yielded very low Pearson correlation coefficients and statistically insignificant p-values in the Mann-Whitney tests. For instance, the overall absurdity level, despite being \textit{High} in the majority of videos, had no discernible statistical relationship with popularity (p=0.2937).

In conclusion, the image-level analysis indicates that while the kinetic and rhythmic aspects of a video's construction are influential, their effect on popularity is not universal. The visual dynamism, characterized by rapid cutting and a fast overall tempo, appears to be a powerful element in capturing and retaining audience attention, but this trend is highly dependent on the cultural context of the video. Our data shows this relationship is strong for Romanian-language content but non-existent for Italian-language content, suggesting that the preference for such a style is culturally specific rather than a general driver of engagement.

\section{Manual analysis}

One of the  most striking brain rot characteristic revealed by the manual scrutiny of all the brain rots represents their core element: the dissonance between the tone, the music and the content. The tone is neutral - an AI generated voice- but the text is often deviant, triggering some extreme emotions. There is no harmony or coherence whatsoever between text, speech, music and image. Another notable trait is the unsettling blend of baby talk, nursery rhymes, and children inappropriate content (topics) and language (slang, pejorative jargon, NSFW words).

Overall, the brain rots seem a good example of E(xtended)-creativity, rather than of F(ixed)-creativity, since they tend to break the rules and not only to create new content using existing ones \cite{dinu-florescu-2025-testing}.

The manual analysis of the textual content revealed the prevalent topics and characters used in both languages. 
Some Romanian brain rot content includes political propaganda related to the presidential elections, often carrying extremist nationalist undertones. This suggests that such content goes beyond seemingly harmless absurdist humor and may serve as manipulative material. 

The Romanian characters are inspired from the Romanian folklore, such as the Balaur (a dragon-like creature), Morcoveață (a carrot shaped boy inspired by Jules Renard’s Poil de Carotte), from Romanian historical figures such as rulers or poets (Mihai Viteazul, Ion Creangă, Mihai Eminescu), or from global pop culture such as Hatsune Miku, Sonic the Hedgehog, or Disney characters, portrayed in explicit real-life situations like relationship with siblings or modern dating.

The topics in Italian brain rot also revolve around daily routine and politics, with celebrity characters such as  Volodimir Zelensky or Emmanuel Macron, portrayed in funny and ironical ways, frequently with misspelled names in order to undermine their authority and to de-glamorize them. There are also very popular characters like Ballerina Cappuccina, a tragic but very graceful figure featuring sometimes grotesque ballet poses, the ironical and dreamy Skeletro, with emo fragility and self deprecating depression and out dated romanticism.

The music is mostly depressive both in popular and unpopular Italian or Romanian brain rot, written in minor tonality which gives to the brain rots a serious, sad and dark sounding. We identified with the Shazam tool a variety of musical pieces, from classical music by Frederic Chopin or Max Richter, through very popular tunes by Ennio Morricone or Bobby McFerrin all the way to some trap and rap pieces. Most of the titles of these pieces include the words \textit{spooky}, \textit{scary}, \textit{depressive}. We noticed that for the popular category the music is slightly less dark than for the unpopular. 
Italian popular brain rots showcase a wide array of soundtracks from various genres, while unpopular ones typically feature the same three most used tracks used by viral brain rots. At the same time, the Romanian unpopular samples contain diverse soundtracks from obscure sources and underground artists. Conversely, a single track is present in more than half of all Romanian popular brain rots. This suggests that more effort was required at the beginning of this trend in Italy, but after gaining notoriety, the key to viral clips was to exploit the same soundtracks as established by the initial wave, making them more recognizable. 

The visuals are overloaded, kitschy, anti-narrative, absurd, over-sized, ironic, cynical, and, in a way, self-destructive.  The popular ones are more animated, present more complex and colorful imagery, the cutting is obviously more dynamic, full of narrative.


\section{Conclusion}

This study conducted a multi-modal analysis of Italian and Romanian brain rot, seeking to identify the factors driving popularity and to map the cultural and linguistic differences between the two languages. Our findings show that the popularity of these memes is not primarily determined by their textual content. Features like sentiment, narrative absurdity, or rhyme density showed no significant link to a video's success. The same can be said about standard audio features and speech sentiment. Instead, the analysis revealed that popularity is strongly correlated with visual dynamics, specifically a faster cutting speed and overall pace (for Romanian language).

The research also uncovered clear distinctions between the Italian and Romanian versions of brain rot. Italian brain rots were textually more negative, more unpredictable, and used rhyme more frequently. Acoustically, their vocals were characterized by greater melodic range and loudness. Conversely, Romanian content was more neutral, while its vocals were spectrally brighter and showed more erratic pitch changes. These differences extend to thematic content, with each language favoring culturally specific characters and references.

As this is a global phenomenon, we plan to extend this study to other Romance languages in future work. Regarding the small sample size, we intend to include samples that would not fit in either popular or unpopular categories, thus rephrasing the popularity aspect as a continuous problem rather than as binary classification.

In essence, the success of brain rot appears to depend on a combination of universal and culturally-specific elements. While fast-paced, dynamic visuals serve as a universal driver for engagement, the content itself is distinctly shaped by the linguistic, acoustic, and thematic norms of its target culture. What makes the genre unique is the strange mix of message, image and sound.

\section{Limitations}
One limitation of this study is that the dataset, while carefully curated, is relatively small and manually collected, focusing only on Italian and Romanian content. This may limit the generalization possibilities of our findings to brain rot in other languages or on a larger scale. Secondly, our analysis identifies strong correlations, such as the link between cutting speed and popularity, but does not establish causation. Other confounding factors not examined here may be at play. Finally, brain rot is a rapidly evolving digital phenomenon, and the features that define it today may change over time, potentially dating our specific observations. 



\begin{acknowledgments}
This research is supported by the project “Romanian
Hub for Artificial Intelligence - HRIA”,
Smart Growth, Digitization and Financial Instruments
Program, 2021-2027, MySMIS no. 334906
and by CNCS/CCCDI UEFISCDI, SiRoLa project,
PN-IV-P1-PCE-2023-1701, within PNCDI IV, Romania.
\end{acknowledgments}

\bibliography{sample-ceur}

\begin{thebibliography}{18}
\expandafter\ifx\csname natexlab\endcsname\relax\def\natexlab#1{#1}\fi
\providecommand{\url}[1]{\texttt{#1}}
\providecommand{\href}[2]{#2}
\providecommand{\path}[1]{#1}
\providecommand{\DOIprefix}{doi:}
\providecommand{\ArXivprefix}{arXiv:}
\providecommand{\URLprefix}{URL: }
\providecommand{\Pubmedprefix}{pmid:}
\providecommand{\doi}[1]{\href{http://dx.doi.org/#1}{\path{#1}}}
\providecommand{\Pubmed}[1]{\href{pmid:#1}{\path{#1}}}
\providecommand{\bibinfo}[2]{#2}
\ifx\xfnm\relax \def\xfnm[#1]{\unskip,\space#1}\fi
\bibitem[{Thoreau(2011)}]{thoreau2011walden}
\bibinfo{author}{H.~Thoreau}, \bibinfo{title}{Walden},
  \bibinfo{publisher}{Mercer University Press}, \bibinfo{year}{2011}.
\bibitem[{Barthes et~al.(1977)Barthes, Heath, and Dove}]{barthes1977death}
\bibinfo{author}{R.~Barthes}, \bibinfo{author}{S.~Heath},
  \bibinfo{author}{M.~Dove}, \bibinfo{title}{The Death of the Author},
  \bibinfo{year}{1977}.
\bibitem[{Yousef et~al.(2025)Yousef, Alshamy, Tlili, and
  Metwally}]{brainsci15030283}
\bibinfo{author}{A.~M.~F. Yousef}, \bibinfo{author}{A.~Alshamy},
  \bibinfo{author}{A.~Tlili}, \bibinfo{author}{A.~H.~S. Metwally},
\newblock \bibinfo{title}{Demystifying the new dilemma of brain rot in the
  digital era: A review},
\newblock \bibinfo{journal}{Brain Sciences} \bibinfo{volume}{15}
  (\bibinfo{year}{2025}). \URLprefix
  \url{https://www.mdpi.com/2076-3425/15/3/283}.
  \DOIprefix\doi{10.3390/brainsci15030283}.
\bibitem[{Satici et~al.(2023)Satici, Tekin, Deniz, and Satici}]{Satici2023}
\bibinfo{author}{S.~A. Satici}, \bibinfo{author}{E.~G. Tekin},
  \bibinfo{author}{M.~E. Deniz}, \bibinfo{author}{B.~Satici},
\newblock \bibinfo{title}{Doomscrolling scale: its association with personality
  traits, psychological distress, social media use, and wellbeing},
\newblock \bibinfo{journal}{Applied Research in Quality of Life}
  \bibinfo{volume}{18} (\bibinfo{year}{2023}) \bibinfo{pages}{833--847}.
  \URLprefix \url{https://doi.org/10.1007/s11482-022-10110-7}.
  \DOIprefix\doi{10.1007/s11482-022-10110-7}.
\bibitem[{Pandiani et~al.(2024)Pandiani, Sang, and
  Ceolin}]{pandiani2024toxicmemessurveycomputational}
\bibinfo{author}{D.~S.~M. Pandiani}, \bibinfo{author}{E.~T.~K. Sang},
  \bibinfo{author}{D.~Ceolin}, \bibinfo{title}{Toxic memes: A survey of
  computational perspectives on the detection and explanation of meme
  toxicities}, \bibinfo{year}{2024}. \URLprefix
  \url{https://arxiv.org/abs/2406.07353}.
  \href{http://arxiv.org/abs/2406.07353}{{\tt arXiv:2406.07353}}.
\bibitem[{Soleymani et~al.(2017)Soleymani, Garcia, Jou, Schuller, Chang, and
  Pantic}]{SOLEYMANI20173}
\bibinfo{author}{M.~Soleymani}, \bibinfo{author}{D.~Garcia},
  \bibinfo{author}{B.~Jou}, \bibinfo{author}{B.~Schuller},
  \bibinfo{author}{S.-F. Chang}, \bibinfo{author}{M.~Pantic},
\newblock \bibinfo{title}{A survey of multimodal sentiment analysis},
\newblock \bibinfo{journal}{Image and Vision Computing} \bibinfo{volume}{65}
  (\bibinfo{year}{2017}) \bibinfo{pages}{3--14}. \URLprefix
  \url{https://www.sciencedirect.com/science/article/pii/S0262885617301191}.
  \DOIprefix\doi{https://doi.org/10.1016/j.imavis.2017.08.003},
  \bibinfo{note}{multimodal Sentiment Analysis and Mining in the Wild Image and
  Vision Computing}.
\bibitem[{Gandhi et~al.(2023)Gandhi, Adhvaryu, Poria, Cambria, and
  Hussain}]{GANDHI2023424}
\bibinfo{author}{A.~Gandhi}, \bibinfo{author}{K.~Adhvaryu},
  \bibinfo{author}{S.~Poria}, \bibinfo{author}{E.~Cambria},
  \bibinfo{author}{A.~Hussain},
\newblock \bibinfo{title}{Multimodal sentiment analysis: A systematic review of
  history, datasets, multimodal fusion methods, applications, challenges and
  future directions},
\newblock \bibinfo{journal}{Information Fusion} \bibinfo{volume}{91}
  (\bibinfo{year}{2023}) \bibinfo{pages}{424--444}. \URLprefix
  \url{https://www.sciencedirect.com/science/article/pii/S1566253522001634}.
  \DOIprefix\doi{https://doi.org/10.1016/j.inffus.2022.09.025}.
\bibitem[{Kiela et~al.(2021)Kiela, Firooz, Mohan, Goswami, Singh, Ringshia, and
  Testuggine}]{kiela2021hatefulmemeschallengedetecting}
\bibinfo{author}{D.~Kiela}, \bibinfo{author}{H.~Firooz},
  \bibinfo{author}{A.~Mohan}, \bibinfo{author}{V.~Goswami},
  \bibinfo{author}{A.~Singh}, \bibinfo{author}{P.~Ringshia},
  \bibinfo{author}{D.~Testuggine}, \bibinfo{title}{The hateful memes challenge:
  Detecting hate speech in multimodal memes}, \bibinfo{year}{2021}. \URLprefix
  \url{https://arxiv.org/abs/2005.04790}.
  \href{http://arxiv.org/abs/2005.04790}{{\tt arXiv:2005.04790}}.
\bibitem[{Lin et~al.(2024)Lin, Luo, Gao, Ma, Wang, and
  Yang}]{lin2024explainableharmfulmemedetection}
\bibinfo{author}{H.~Lin}, \bibinfo{author}{Z.~Luo}, \bibinfo{author}{W.~Gao},
  \bibinfo{author}{J.~Ma}, \bibinfo{author}{B.~Wang},
  \bibinfo{author}{R.~Yang}, \bibinfo{title}{Towards explainable harmful meme
  detection through multimodal debate between large language models},
  \bibinfo{year}{2024}. \URLprefix \url{https://arxiv.org/abs/2401.13298}.
  \href{http://arxiv.org/abs/2401.13298}{{\tt arXiv:2401.13298}}.
\bibitem[{Bloem and Ilievski(2025)}]{bloem2025clusteringinternetmemestemplate}
\bibinfo{author}{T.~Bloem}, \bibinfo{author}{F.~Ilievski},
  \bibinfo{title}{Clustering internet memes through template matching and
  multi-dimensional similarity}, \bibinfo{year}{2025}. \URLprefix
  \url{https://arxiv.org/abs/2505.00056}.
  \href{http://arxiv.org/abs/2505.00056}{{\tt arXiv:2505.00056}}.
\bibitem[{Barbieri et~al.(2022)Barbieri, Espinosa~Anke, and
  Camacho-Collados}]{barbieri-etal-2022-xlm}
\bibinfo{author}{F.~Barbieri}, \bibinfo{author}{L.~Espinosa~Anke},
  \bibinfo{author}{J.~Camacho-Collados},
\newblock \bibinfo{title}{{XLM}-{T}: Multilingual language models in {T}witter
  for sentiment analysis and beyond},
\newblock in: \bibinfo{booktitle}{Proceedings of the Thirteenth Language
  Resources and Evaluation Conference}, \bibinfo{publisher}{European Language
  Resources Association}, \bibinfo{address}{Marseille, France},
  \bibinfo{year}{2022}, pp. \bibinfo{pages}{258--266}. \URLprefix
  \url{https://aclanthology.org/2022.lrec-1.27}.
\bibitem[{Reimers and Gurevych(2019)}]{reimers-gurevych-2019-sentence}
\bibinfo{author}{N.~Reimers}, \bibinfo{author}{I.~Gurevych},
\newblock \bibinfo{title}{Sentence-{BERT}: Sentence embeddings using {S}iamese
  {BERT}-networks},
\newblock in: \bibinfo{editor}{K.~Inui}, \bibinfo{editor}{J.~Jiang},
  \bibinfo{editor}{V.~Ng}, \bibinfo{editor}{X.~Wan} (Eds.),
  \bibinfo{booktitle}{Proceedings of the 2019 Conference on Empirical Methods
  in Natural Language Processing and the 9th International Joint Conference on
  Natural Language Processing (EMNLP-IJCNLP)}, \bibinfo{publisher}{Association
  for Computational Linguistics}, \bibinfo{address}{Hong Kong, China},
  \bibinfo{year}{2019}, pp. \bibinfo{pages}{3982--3992}. \URLprefix
  \url{https://aclanthology.org/D19-1410/}.
  \DOIprefix\doi{10.18653/v1/D19-1410}.
\bibitem[{Radford et~al.(2019)Radford, Wu, Child, Luan, Amodei, Sutskever
  et~al.}]{radford2019language}
\bibinfo{author}{A.~Radford}, \bibinfo{author}{J.~Wu},
  \bibinfo{author}{R.~Child}, \bibinfo{author}{D.~Luan},
  \bibinfo{author}{D.~Amodei}, \bibinfo{author}{I.~Sutskever}, et~al.,
\newblock \bibinfo{title}{Language models are unsupervised multitask learners},
\newblock \bibinfo{journal}{OpenAI blog}  (\bibinfo{year}{2019}). \URLprefix
  \url{https://cdn.openai.com/better-language-models/language_models_are_unsupervised_multitask_learners.pdf}.
\bibitem[{Rouard et~al.(2023)Rouard, Massa, and Défossez}]{rouard2023hybrid}
\bibinfo{author}{S.~Rouard}, \bibinfo{author}{F.~Massa},
  \bibinfo{author}{A.~Défossez},
\newblock \bibinfo{title}{Hybrid transformers for music source separation},
\newblock in: \bibinfo{booktitle}{ICASSP 2023 - 2023 IEEE International
  Conference on Acoustics, Speech and Signal Processing (ICASSP)},
  \bibinfo{year}{2023}, pp. \bibinfo{pages}{1--5}.
  \DOIprefix\doi{10.1109/ICASSP49357.2023.10096956}.
\bibitem[{McFee et~al.(2025)McFee, McVicar, Faronbi, Roman, Gover
  et~al.}]{mcfee_2025_15006942}
\bibinfo{author}{B.~McFee}, \bibinfo{author}{M.~McVicar},
  \bibinfo{author}{D.~Faronbi}, \bibinfo{author}{I.~Roman},
  \bibinfo{author}{M.~Gover}, et~al., \bibinfo{title}{Librosa},
  \bibinfo{year}{2025}. \URLprefix
  \url{https://doi.org/10.5281/zenodo.15006942}.
  \DOIprefix\doi{10.5281/zenodo.15006942}.
\bibitem[{Benjamini and Hochberg(2018)}]{benjamini-95-controlling}
\bibinfo{author}{Y.~Benjamini}, \bibinfo{author}{Y.~Hochberg},
\newblock \bibinfo{title}{Controlling the false discovery rate: A practical and
  powerful approach to multiple testing},
\newblock \bibinfo{journal}{Journal of the Royal Statistical Society: Series B
  (Methodological)} \bibinfo{volume}{57} (\bibinfo{year}{2018})
  \bibinfo{pages}{289--300}. \URLprefix
  \url{https://doi.org/10.1111/j.2517-6161.1995.tb02031.x}.
  \DOIprefix\doi{10.1111/j.2517-6161.1995.tb02031.x}.
\bibitem[{Yang et~al.(2021)Yang, Chi, Chuang, Lai, Lakhotia, Lin, Liu, Shi,
  Chang, Lin et~al.}]{yang2021superb}
\bibinfo{author}{S.-w. Yang}, \bibinfo{author}{P.-H. Chi},
  \bibinfo{author}{Y.-S. Chuang}, \bibinfo{author}{C.-I.~J. Lai},
  \bibinfo{author}{K.~Lakhotia}, \bibinfo{author}{Y.~Y. Lin},
  \bibinfo{author}{A.~T. Liu}, \bibinfo{author}{J.~Shi},
  \bibinfo{author}{X.~Chang}, \bibinfo{author}{G.-T. Lin}, et~al.,
  \bibinfo{title}{{SUPERB}: Speech processing {U}niversal {PER}formance
  {B}enchmark}, \bibinfo{year}{2021}.
  \href{http://arxiv.org/abs/2105.01051}{{\tt arXiv:2105.01051}}.
\bibitem[{Dinu and Florescu(2025)}]{dinu-florescu-2025-testing}
\bibinfo{author}{A.~Dinu}, \bibinfo{author}{A.-M. Florescu},
\newblock \bibinfo{title}{Testing language creativity of large language models
  and humans},
\newblock in: \bibinfo{editor}{M.~H{\"a}m{\"a}l{\"a}inen},
  \bibinfo{editor}{E.~{\"O}hman}, \bibinfo{editor}{Y.~Bizzoni},
  \bibinfo{editor}{S.~Miyagawa}, \bibinfo{editor}{K.~Alnajjar} (Eds.),
  \bibinfo{booktitle}{Proceedings of the 5th International Conference on
  Natural Language Processing for Digital Humanities},
  \bibinfo{publisher}{Association for Computational Linguistics},
  \bibinfo{address}{Albuquerque, USA}, \bibinfo{year}{2025}, pp.
  \bibinfo{pages}{426--436}. \URLprefix
  \url{https://aclanthology.org/2025.nlp4dh-1.37/}.
  \DOIprefix\doi{10.18653/v1/2025.nlp4dh-1.37}.

\end{thebibliography}


\onecolumn
\appendix
\clearpage

\section*{Appendix - Supplementary statistical tests tables}

\begin{table*}[htbp!]
\tiny
  \caption{Statistical tests for Sentiment Analysis of the texts}
  \label{tab:statsSentimentsText}
  \begin{tabular}{lllll}
    \toprule
   pair & MANOVA & Mann-Whitney for neg. & Mann-Whitney for neu. & Mann-Whitney for pos.\\
    \midrule
     Ro pop. - Ro unpop. & $\lambda$ = 0.99, F(3, 116) = 0.55, p = 0.65 & p = 0.79 (M: 0.38 vs. 0.36) & p = 0.57 (M: 0.39 vs. 0.40) & p = 0.990 (M: 0.24 vs. 0.23)\\
    It pop. - It unpop. & $\lambda$ = 0.97, F(3, 116) = 1.15, p = 0.33 & p = 0.36 (M: 0.62 vs. 0.67) & p = 0.54 (M: 0.19 vs. 0.17) & p = 0.39 (M: 0.19 vs. 0.16) \\
   It - Ro & $\lambda$ = 0.40, F(3, 116) = 56.74, p < 0.001 & p < 0.0001 (M: 0.38 vs. 0.16) & p < 0.0001 (M: 0.39 vs. 0.17) & p < 0.0001 (M: 0.67 vs. 0.24)  \\
       \bottomrule
\end{tabular}


\tiny
\centering
\caption{Statistical tests for Semantic Similarity and Perplexity of the texts}
\label{tab:stats_combinedSemanticSimilarityAndPerplexity}
\begin{tabular}{lll @{\hskip 1cm} lll}
\toprule
\multicolumn{3}{c}{\textbf{Semantic Similarity}} & \multicolumn{3}{c}{\textbf{Perplexity}} \\
\cmidrule(lr){1-3} \cmidrule(lr){4-6}
\textbf{Pair} & \textbf{ANOVA} & \textbf{Mann-Whitney U Test} &
\textbf{Pair} & \textbf{ANOVA} & \textbf{Mann-Whitney U Test} \\
\midrule
Ro pop. - Ro unpop. & F(1, 118) = 2.35, p = 0.128 & p = 0.068 (M = 0.58 vs. 0.57) &
Ro pop. - Ro unpop. & F(1, 118) = 0.17, p = 0.684 & p = 0.927 (M = 168.59 vs. 159.22)\\
It pop. - It unpop. & F(1, 118) = 2.38, p = 0.126 & p = 0.182 (M = 0.58 vs. 0.59) &
It pop. - It unpop. & F(1, 118) = 0.78, p = 0.380 & p = 0.821 (M = 228.92 vs. 208.20)\\
It - Ro & F(1, 238) = 1.88, p = 0.172 & p = 0.0614 (M = 0.59 vs. 0.58) & It - Ro & F(1, 238) = 11.19, p < 0.001 & p < .0001 (M = 218.74 vs. 163.91)\\
\bottomrule
\end{tabular}


\tiny
  \caption{Statistical Tests for Rhyme Coefficient}
  \label{tab:stats_rhyme}
  \begin{tabular}{lcl}
    \toprule
    Pair & Descriptive Statistics & Mann-Whitney\\
    \midrule
    Ro pop. - Ro unpop. & M = 0.195 vs. 0.161 & p = 0.135\\
    It pop. - It unpop. & M = 0.215 vs. 0.228 & p = 0.688\\
    It - Ro & M = 0.222 vs. 0.178 & p = 0.0001\\
    \bottomrule
  \end{tabular}


  \centering
  \tiny
      \caption{Comparison of key audio features between Italian and Romanian brain rot, after FDR correction that differ at q<0.05}
  \label{tab:italian_vs_romanian}
    \begin{tabular}{l r r l}
      \toprule
      Feature                            & Italian & Romanian & q-value \\
      \midrule



      
      var\_f\textsubscript{0} (Hz\textsuperscript{2})                & 4570.8  & 2234.0   & $9.90\times10^{-6}$   \\
      range\_f\textsubscript{0} (Hz)                  & 162.0   & 119.1    & $3.50\times10^{-6}$   \\
      entropy\_f\textsubscript{0}\_slope (bits)  & 1.79    & 2.05     & $1.96\times10^{-4}$   \\
      rms\_mean                          & 0.141   & 0.123    & $1.39\times10^{-2}$   \\
      spec\_centroid\_mean (Hz)          & 1961    & 2161     & $1.39\times10^{-8}$   \\
      spec\_centroid\_var (Hz\textsuperscript{2})       & 1.24e6  & 1.67e6   & $5.99\times10^{-18}$  \\
      spec\_bandwidth\_mean (Hz)         & 1936    & 2059     & $3.05\times10^{-8}$   \\
      spec\_bandwidth\_var (Hz\textsuperscript{2})      & 2.45e5  & 2.84e5   & $4.71\times10^{-6}$   \\
      spec\_rolloff\_mean (Hz)           & 3625    & 4000     & $1.78\times10^{-7}$   \\
      spec\_rolloff\_var (Hz\textsuperscript{2})        & 3.50e6  & 4.45e6   & $4.42\times10^{-15}$  \\
      zcr\_mean                          & 0.0975  & 0.1098   & $6.58\times10^{-7}$   \\
      zcr\_var                           & 0.00868 & 0.01162  & $2.46\times10^{-12}$  \\
      flux\_var                          & 8.84    & 9.73     & $6.64\times10^{-4}$   \\
      mfcc\_1\_mean                      & -197.4  & -220.7   & $4.75\times10^{-3}$   \\
      mfcc\_1\_var                       & 1.79e4  & 2.30e4   & $7.29\times10^{-10}$  \\
      $\cdots$ & $\cdots$ & $\cdots$ & $\cdots$ \\
      \bottomrule
    \end{tabular}
      \hspace{5em}
    \begin{tabular}{l r r l}
      \toprule
      Feature                            & Italian & Romanian & q-value \\
      \midrule      
      mfcc\_2\_mean                      & 93.7    & 86.4     & $1.60\times10^{-4}$   \\
      mfcc\_2\_var                       & 3.73e3  & 4.24e3   & $1.23\times10^{-4}$   \\
      mfcc\_4\_mean                      & 25.3    & 20.7     & $4.42\times10^{-5}$   \\
      mfcc\_4\_var                       & 892     & 790      & $3.12\times10^{-5}$   \\
      mfcc\_5\_var                       & 627     & 499      & $5.61\times10^{-11}$  \\
      mfcc\_6\_mean                      & -4.48   & -2.52    & $1.17\times10^{-2}$   \\
      mfcc\_7\_mean                      & -11.42  & -8.57    & $4.75\times10^{-6}$   \\
      mfcc\_7\_var                       & 318     & 271      & $1.38\times10^{-10}$  \\
      mfcc\_9\_mean                      & -13.44  & -11.34   & $6.97\times10^{-7}$   \\
      mfcc\_9\_var                       & 218     & 210      & $2.70\times10^{-2}$   \\
      mfcc\_10\_mean                     & -9.10   & -7.59    & $5.97\times10^{-3}$   \\
      mfcc\_11\_mean                     & -5.82   & -4.92    & $8.36\times10^{-4}$   \\
      mfcc\_11\_var                      & 203     & 167      & $4.71\times10^{-10}$  \\
      mfcc\_12\_mean                     & -3.49   & -2.40    & $8.03\times10^{-3}$   \\
      mfcc\_12\_var                      & 112     & 98       & $3.91\times10^{-6}$   \\
      mfcc\_13\_var                      & 105     & 101      & $2.24\times10^{-3}$   \\
      \bottomrule
    \end{tabular}


  \tiny
    \caption{Raw p < 0.05 comparisons of audio features between all popular and unpopular brain rots within Italian and Romanian corpora}
  \label{tab:pop_unpop_features}
  \begin{tabular}{l l r r c}
    \toprule
    Language  & Feature & Popular & Unpopular & p‐value \\
    \midrule
    \textbf{Italian}  & var\_f0 (Hz\textsuperscript{2})            & 6657   & 2484   & 0.0160  \\
                      & range\_f0 (Hz)                &  189   &  135   & 0.0028  \\
    \midrule
    \textbf{Romanian} & syllable\_rate (onsets/s)     & 4.29   & 4.09   & 0.0260  \\
                      & mfcc\_9\_var                  & 202    & 217    & 0.0062  \\
                      & mfcc\_11\_mean                & –5.65  & –4.18  & 0.0300  \\
    \bottomrule
  \end{tabular}


\tiny
\centering
\caption{Image attributes that might have been a good indicator of video popularity}
\label{table:video_analysis_full}
\begin{tabular}{l c c c}
\toprule
\textbf{Attribute} & \textbf{Pearson Correlation} & \textbf{Mann-Whitney P-value} & \textbf{Significant (<0.05)} \\
\midrule 
Guessed Popularity (Visual) & 0.0138 & 0.6001 & No \\
Cutting Speed & 0.1712 & 0.0162 & \textbf{Yes} \\
Overall Pacing & 0.1226 & 0.0492 & \textbf{Yes} \\
Subject Movement & 0.1053 & 0.0858 & No \\
Narrative Logic & 0.0711 & 0.2485 & No \\
Non-Sequitur Visuals & -0.0703 & 0.2785 & No \\
Overall Absurdity Level & 0.0584 & 0.2937 & No \\
\bottomrule
\end{tabular}
\end{table*}










\end{document}